\pdfoutput=1

\documentclass[11pt]{article}

\usepackage[]{EMNLP2023}

\usepackage{times}
\usepackage{latexsym}

\usepackage[T1]{fontenc}

\usepackage[utf8]{inputenc}

\usepackage{microtype}

\usepackage{inconsolata}

\usepackage{bm}
\usepackage{amsmath}
\usepackage{url}
\usepackage{graphicx}
\usepackage{float} 
\usepackage{multirow}
\usepackage{makecell}
\usepackage{xpinyin}
\usepackage{enumerate}

\usepackage{xcolor}
\usepackage{framed}
\usepackage{color}
\usepackage{bbding}
\usepackage{soul}
\usepackage{subcaption}
\usepackage{colortbl}
\newcolumntype{L}[1]{>{\raggedright\let\newline\\\arraybackslash\hspace{0pt}}m{#1}}
\setul{2pt}{2pt}
\usepackage{booktabs}
\usepackage{amsmath,amsfonts}
\usepackage[utf8]{inputenc}
\usepackage{cleveref}

\usepackage{ulem}
\usepackage{diagbox}
\crefname{section}{§}{§§}
\Crefname{section}{§}{§§}
\usepackage{enumitem}
\setenumerate[1]{itemsep=0pt,partopsep=0pt,parsep=\parskip,topsep=5pt}
\setitemize[1]{itemsep=0pt,partopsep=0pt,parsep=\parskip,topsep=5pt}
\setdescription{itemsep=0pt,partopsep=0pt,parsep=\parskip,topsep=5pt}
\usepackage{arydshln}

\usepackage[linesnumbered,ruled,vlined]{algorithm2e}

\definecolor{ggray}{HTML}{E7E6E6}
\definecolor{ggreen}{HTML}{d4e7cf}

\title{Dual Knowledge Distillation for Many-to-Many Multimodal Summarization}
\title{D$^2$TV: Dual Knowledge Distillation and Target-oriented Vision Modeling for Many-to-Many Multimodal Summarization}

\author{
  Yunlong Liang\textsuperscript{1}\thanks{ \ \ Work was done when Liang and Wang was interning at Pattern Recognition Center, WeChat AI, Tencent Inc, China.}  , 
  Fandong Meng\textsuperscript{2},  
  Jiaan Wang\textsuperscript{3}, 
  \textbf{Jinan Xu}\textsuperscript{1}\thanks{ \ \ Jinan Xu is the corresponding author.}  ,
  \textbf{Yufeng Chen}\textsuperscript{1}
   and \textbf{Jie Zhou}\textsuperscript{2}\\
  \textsuperscript{1}Beijing Key Lab of Traffic Data Analysis and Mining, \\Beijing Jiaotong University, Beijing, China \\
  \textsuperscript{2}Pattern Recognition Center, WeChat AI, Tencent Inc, China \\
  \textsuperscript{3}School of Computer Science and Technology, Soochow University, Suzhou, China\\
  \texttt{\{yunlongliang,jaxu,chenyf\}@bjtu.edu.cn} \\
  \texttt{\{fandongmeng,withtomzhou\}@tencent.com}\ \ \ \ \ \texttt{jawang.nlp@gmail.com}\\
}

\begin{document}
\maketitle
\begin{abstract}

Many-to-many multimodal summarization (M$^3$S) task aims to generate summaries in any language with document inputs in any language and the corresponding image sequence, which essentially comprises multimodal monolingual summarization (MMS) and multimodal cross-lingual summarization (MXLS) tasks. Although much work has been devoted to either MMS or MXLS and has obtained increasing attention in recent years, little research pays attention to the M$^3$S task. Besides, existing studies mainly focus on 1) utilizing MMS to enhance MXLS via knowledge distillation without considering the performance of MMS or 2) improving MMS models by filtering summary-unrelated visual features with implicit learning or explicitly complex training objectives. In this paper, we first introduce a general and practical task, \emph{i.e.}, M$^3$S. Further, we propose a dual knowledge distillation and target-oriented vision modeling framework for the M$^3$S task. Specifically, the dual knowledge distillation method guarantees that the knowledge of MMS and MXLS can be transferred to each other and thus mutually prompt both of them. To offer target-oriented visual features, a simple yet effective target-oriented contrastive objective is designed and responsible for discarding needless visual information. Extensive experiments on the many-to-many setting show the effectiveness of the proposed approach. Additionally, we will contribute a many-to-many multimodal summarization ({\fontfamily{lmtt}\selectfont M$^3$Sum}) dataset.\footnote{The code and data are publicly available at \url{https://github.com/XL2248/D2TV}.}

\end{abstract}

\section{Introduction}

\begin{figure}[t]
    \centering
    \includegraphics[width=0.98\columnwidth]{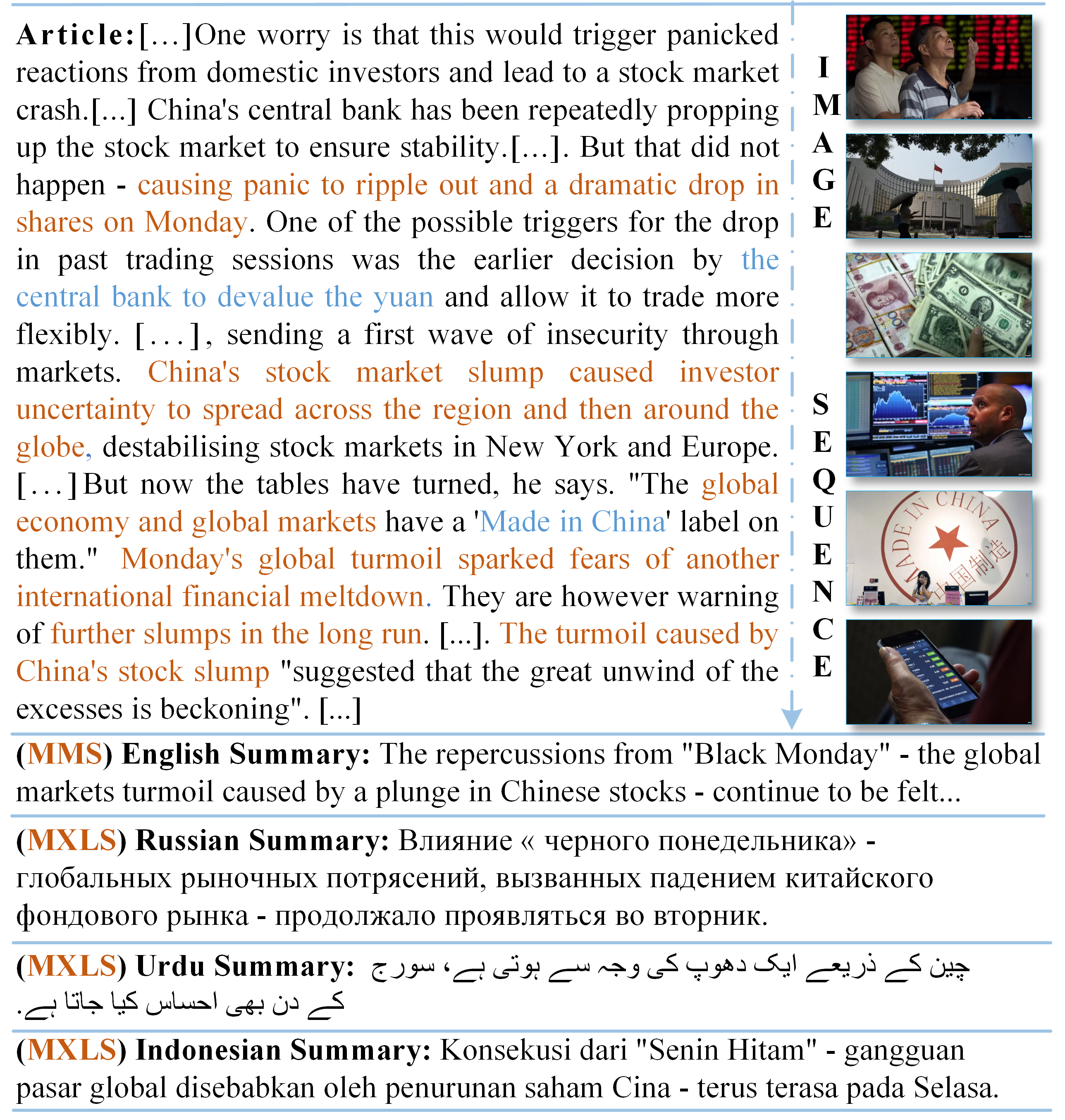}
    \caption{An example of our {\fontfamily{lmtt}\selectfont M$^3$Sum} dataset. Inputs: an article and corresponding image sequence; Output: summaries in different languages. MMS: the summary in the same language as the input article; MXLS: the summary in a different language from the input article. The M$^3$S setting covers both MMS and MXLS.}
    \label{fig:evc}\vspace{-10pt}
\end{figure}

Given a document input in the source language (\emph{e.g.}, English) and its corresponding image sequence, multimodal monolingual summarization (MMS) aims to generate a summary in the same language (\emph{i.e.}, English) while the goal of multimodal cross-lingual summarization (MXLS) is to produce a summary in a different language (\emph{e.g.}, Chinese). With the rapid increase of multimedia data, the MMS~\cite{5711541,6527322,1221239,li-etal-2017-multi,li2018multi,sanabria18how2,zhu-etal-2018-msmo,chen-zhuge-2018-abstractive,li2020aspect,fu-etal-2021-mm,zhao-etal-2022-jddc} and MXLS~\cite{liu-etal-2022-assist} tasks have attracted much attention in the research community because both tasks can help users quickly master the core idea from the cumbersome multimodal data. Essentially, the many-to-many multimodal summarization (M$^3$S) consists of MMS and MXLS tasks, which generate summaries in any language given the multimodal inputs (in any language), as~\autoref{fig:evc} shows. Intuitively, the many-to-many setup should be more general and practical for its application in the multilingual and multimodal world~\cite{wang2023towards}. 

In the literature, although plenty of studies have been carried out on MMS or MXLS, there is only one study that involves both of them, \emph{i.e.}, ~\citet{liu-etal-2022-assist} devise a triple-stage training framework and distill the knowledge from MMS to enhance MXLS while ignoring the performance of MMS. Despite their effectiveness on MXLS, to our knowledge, little research attention has been paid to simultaneously supporting both MMS and MXLS tasks and prompting both of them. Besides, the visual features generally include noise which is summary-unrelated. Thus the remaining work mainly focuses on improving MMS models by filtering these noises with (a) implicit learning or (b) complex training objectives. For (a), researchers design various fusion methods to effectively model the interactions between textual articles and visual features~\cite{liu-etal-2020-multistage,yu-etal-2021-vision,palaskar-etal-2019-multimodal,zhang2021hierarchical}. For (b), to explicitly filter needless visual information, ~\citet{https://doi.org/10.48550/arxiv.2212.07672} present two well-designed auxiliary tasks, \emph{i.e.}, vision to summary and masked image modeling. Albeit effective, implicit learning via the MMS objective may limit the potential of visual features, and explicit training objectives are complex and time-consuming to be trained and applied in the real world.

To address these issues, in this paper, we first introduce a more general task, \emph{i.e.}, M$^3$S, which supports both MMS and MXLS tasks. Further, we propose a \textbf{D}ual knowledge \textbf{D}istillation and \textbf{T}arget-oriented \textbf{V}ision enhanced framework, named D$^2$TV, for the new task. Specifically, the dual knowledge distillation approach ensures that the knowledge from MMS can be transferred to MXLS and vice versa, and thus mutually improve both tasks. Furthermore, to discard the summary-unrelated visual information, a target-oriented contrastive objective is devised to directly optimize the visual features. In this way, the model is enhanced to explicitly exploit the summary-oriented visual features, thereby yielding more accurate summaries.

To validate the D$^2$TV framework, we provide a \textbf{M}any-to-\textbf{M}any \textbf{M}ultimodal \textbf{Sum}marization ({\fontfamily{lmtt}\selectfont M$^3$Sum}) benchmark dataset by reorganizing the cross-lingual summarization dataset~\cite{bhattacharjee2022crosssum} and MM-Sum dataset~\cite{https://doi.org/10.48550/arxiv.2212.07672}. The {\fontfamily{lmtt}\selectfont M$^3$Sum} covers 44 languages and thus involves 44*44 language directions. To efficiently evaluate our approach, we randomly select 4 languages (\emph{i.e.}, English, Indonesian, Russian, and Urdu\footnote{Urdu is the low-resource language.}), which consist of 4*4 language directions. We implement our approach grounding on two generative pre-trained language models, \emph{i.e.}, mT5~\cite{xue-etal-2021-mt5} and mBART-50~\cite{tang-etal-2021-multilingual}. Extensive experiments on both backbones show that our model significantly outperforms related methods in terms of ROUGE~\cite{lin-2004-rouge} and BERTScore~\cite{bert-score} scores, demonstrating its effectiveness. The human evaluation further suggests the superiority of our approach. In summary, our main contributions are:

\begin{itemize}[leftmargin=*]

\item To the best of our knowledge, we are the first that introduces the general many-to-many multimodal summarization (M$^3$S) task and contributes a corresponding benchmark dataset. 

\item We propose a dual knowledge distillation and target-oriented vision modeling framework for the M$^3$S task.

\item Experiments on {\fontfamily{lmtt}\selectfont M$^3$Sum} benchmark show that our model builds new state-of-the-art performance, showing the effectiveness of the proposed approach.

\end{itemize}

\section{Background}\label{bg}

\subsection{Problem Formulation}
\label{pf}
Given an input article $\mathcal{X}^{L_1}$$=$$\{x_k^{L_1}\}_{k=1}^{M}$ in language $L_1$ and its corresponding visual features $\mathcal{V}$$=$$\{v_{ij}\}_{i=1,j=1}^{i\leq n,j\leq m}$, where $x_k^{L_1}$ denotes the $k$-th token, and $M$ is the number of tokens in the article, and $v_{ij}$ represents the detected $j$-th object of the $i$-th image ($n$, $m$ is the number of images and detected objects in each image, respectively), the many-to-many multimodal summarization task is defined as:
\begin{equation}\nonumber
\label{eq:ms}
    p(\mathcal{Y}^{L_2}|\mathcal{X}^{L_1}, \mathcal{V}) = \prod_{t=1}^{N}p(y_t^{L_2}|\mathcal{X}^{L_1}, \mathcal{V}, y_{<t}^{L_2}),
\end{equation}
where $y_{<t}^{L_2}$ indicates the tokens before the $t$-th time step of the summary $\mathcal{Y}^{L_2}$$=$$\{y_t^{L_2}\}_{t=1}^{N}$ in language $L_2$ and $N$ is the number of tokens in the summary. The $L_1$ and $L_2$ can be any language.

\subsection{The MMS Model}
\label{MMS}

Following~\citet{yu-etal-2021-vision,https://doi.org/10.48550/arxiv.2212.07672}, the MMS model is an extension of the pre-trained language model (\emph{e.g.}, mT5~\cite{xue-etal-2021-mt5}) based on Transformer architecture~\cite{vaswani2017attention}. As shown in  the left part of~\autoref{fig.2}, it includes four modules: textual encoder, visual encoder, text-vision fusion, and decoder.

\noindent \textbf{Textual Encoder.} The textual encoder consists of $N_e$ stacked layers, where each layer consists of two sub-layers, a multihead self-attention sub-layer ($\mathrm{SelfAttn}$) and a position-wise feed-forward network ($\mathrm{FFN}$) sub-layer:
\begin{equation}
\setlength{\abovedisplayskip}{5pt}
\setlength{\belowdisplayskip}{5pt}
\resizebox{0.89\hsize}{!}{$
\begin{split}
    \mathbf{S}^\ell_{T} &= \mathrm{SelfAttn}(\mathbf{H}^{\ell-1}_{T}) + \mathbf{H}^{\ell-1}_{T},\ \mathbf{S}^\ell_{T} \in \mathbb{R}^{M \times d},\nonumber\\
    \mathbf{H}^\ell_{T} &= \mathrm{FFN}(\mathbf{S}^\ell_{T}) + \mathbf{S}^\ell_{T},\ \mathbf{H}^\ell_{T} \in \mathbb{R}^{M \times d},\nonumber
\end{split}
$}
\end{equation}
where $\mathbf{H}^{\ell-1}_{T}$ and $\mathbf{H}^{\ell}_{T}$ denote the inputs and outputs of the $\ell$-th encoder layer, respectively, and $\mathbf{H}^{0}_{T}$ is initialized as the embedding of input tokens $\mathcal{X}^{L_1}$ and $d$ is the hidden dimension. 

\noindent \textbf{Visual Encoder.} 
Following~\citet{yu-etal-2021-vision,liang2020infusing,LIANG2022103714,liang2022msctd,zhang2021hierarchical,zhang2021unims}, the visual encoder is also the Transformer~\cite{vaswani2017attention} encoder with $N_v$ stacked layers. The difference is the visual inputs. Generally, there is an image sequence to be extracted by the Faster R-CNNs~\cite{NIPS2015_14bfa6bb} pre-trained on Visual Genome~\cite{krishnavisualgenome}. Specifically, for the $i$-th input image, we obtain a set of detected objects from Faster R-CNNs, \emph{i.e.}, $\mathbf{I}_i$ = $\{\mathbf{o}_{i,1}, \mathbf{o}_{i,2}, \mathbf{o}_{i,3}, ..., \mathbf{o}_{i,m}\}$, where $m$ is the number of extracted objects and $\mathbf{o}_{i,*} \in \mathbb{R}^{d_v}$. Each object is captured by a dense feature representation, which can be mapped back to a bounding box / region (\emph{i.e.}, Region-of-Interest (RoI)). Finally, the image sequence is converted to visual features $\mathbf{I}$$=$$\{\mathbf{o}_{ij}\}_{i=1,j=1}^{i\leq n,j\leq m}$. Following~\citet{pmlr-v139-cho21a}, the RoI bounding box coordinates $\mathbf{E}^{box}_{ij}$, image id embedding $\mathbf{E}^{img}_{i}$, and region id embedding $\mathbf{E}^{reg}_{j}$ are added on the visual features to keep the order information of the image sequence:
\begin{equation}\label{input_embed2}\nonumber
\setlength{\abovedisplayskip}{5pt}
\setlength{\belowdisplayskip}{5pt}
\mathbf{v}_{ij} = \mathbf{o}_{ij} + \mathbf{E}^{box}_{ij} + \mathbf{E}^{img}_{i} + \mathbf{E}^{reg}_{j}; i \leq n,j\leq m.
\end{equation}
Then, they are fed into the visual encoder for better modeling the intramodal dynamics and enhancing the vision-specific order information. 
\begin{equation}
\setlength{\abovedisplayskip}{5pt}
\setlength{\belowdisplayskip}{5pt}
\resizebox{0.89\hsize}{!}{$
\begin{split}
    \mathbf{S}^\ell_{V} &= \mathrm{SelfAttn}(\mathbf{H}^{\ell-1}_{V}) + \mathbf{H}^{\ell-1}_{V},\ \mathbf{S}^\ell_{V} \in \mathbb{R}^{|\mathcal{V}| \times d_v},\nonumber\\
    \mathbf{H}^\ell_{V} &= \mathrm{FFN}(\mathbf{S}^\ell_{V}) + \mathbf{S}^\ell_{V},\ \mathbf{H}^\ell_{V} \in \mathbb{R}^{|\mathcal{V}| \times d_v},\nonumber
\end{split}
$}
\end{equation}
where $\mathbf{H}^{\ell-1}_{V}$ and $\mathbf{H}^{\ell}_{V}$ denote the inputs and outputs of the $\ell$-th encoder layer, respectively, and $\mathbf{H}^{0}_{V}$ is initialized as the $\mathbf{Z}$$=$$\{\mathbf{v}_{ij}\}_{i=1,j=1}^{i\leq n,j\leq m}$, and $d_v$ is the hidden dimension. 

\noindent \textbf{Text-Vision Fusion.} Following~\citet{yu-etal-2021-vision}, the visual features are firstly injected by cross-modal multi-head attention ($\mathrm{CrossMAttn}$):
\begin{equation}\nonumber
\setlength{\abovedisplayskip}{5pt}
\setlength{\belowdisplayskip}{5pt}
    \mathbf{M} = \mathrm{CrossMAttn}(\mathbf{Q},\mathbf{K},\mathbf{V}), \ \mathbf{M} \in \mathbb{R}^{M \times d_c}, 
\end{equation}
where $\mathbf{Q}$ are the projected textual features $\mathbf{Q} = \mathbf{H}_{T}^{N_e} \mathbf{W}_q$, $\mathbf{K}$ and $\mathbf{V}$ are the projected visual features with different weights, \emph{i.e.}, $\mathbf{K} = \mathbf{H}_V^{N_v} \mathbf{W}_k$, $\mathbf{V} = \mathbf{H}_V^{N_v} \mathbf{W}_v$, and $ \mathbf{Q} \in \mathbb{R}^{M \times d_c}$, $\mathbf{K}, \mathbf{V} \in \mathbb{R}^{|\mathcal{V}| \times d_c}$, and $d_c$ is the common hidden dimension.

Secondly, a forget gate $\mathbf{G}$ is used to filter redundant and noisy information from the visual features:
\begin{equation}\nonumber
\setlength{\abovedisplayskip}{5pt}
\setlength{\belowdisplayskip}{5pt}
\resizebox{0.89\hsize}{!}{$
\begin{split}
    \mathbf{G} &= \mathrm{Sigmoid}(\mathrm{Concat}(\mathbf{H}_{T}^{N_e}, \mathbf{M}) \mathbf{W}_g + \mathbf{b}_g), \\
    \mathbf{Z}_V &= \mathbf{G} \otimes \mathbf{M}.
\end{split}
$}
\end{equation}

Finally, the vision-guided output $\mathbf{Z}_{T+V}$ is concatenated by $\mathbf{Z}_V$ and textual features $\mathbf{H}_{T}^{N_e}$, and then linearly project it to the original dimension $d$:
\begin{equation}\nonumber
\setlength{\abovedisplayskip}{5pt}
\setlength{\belowdisplayskip}{5pt}
\begin{split}
    \mathbf{Z}_{T+V} &= \mathrm{Concat}(\mathbf{H}_{T}^{N_e}, \mathbf{Z}_V) \mathbf{W}_z + \mathbf{b}_z,
\end{split}
\end{equation}
where $\mathrm{Concat}$ is the concatenation operation and $\mathbf{W}_*$ and $\mathbf{b}_*$ are trainable weights.

\noindent \textbf{Decoder.} The decoder follows a similar architecture but each of $N_d$ decoder layers has an additional multi-head cross-attention ($\mathrm{CrossAttn}$) sub-layer:
\begin{equation}
\setlength{\abovedisplayskip}{5pt}
\setlength{\belowdisplayskip}{5pt}
\resizebox{0.89\hsize}{!}{$
\begin{split}
\label{eq:decoder}
    \mathbf{S}^\ell_{dec} &= \mathrm{SelfAttn}(\mathbf{H}^{\ell-1}_{dec}) + \mathbf{H}^{\ell-1}_{dec},\ \mathbf{S}^{\ell-1}_{dec}\in \mathbb{R}^{N \times d},\\
    \mathbf{C}^\ell_{dec} &= \mathrm{CrossAttn}(\mathbf{S}^{\ell}_{dec}, \mathbf{Z}_{T+V}) + \mathbf{S}^\ell_{dec}, \\
    \mathbf{H}^\ell_{dec} &= \mathrm{FFN}(\mathbf{C}^\ell_{dec}) + \mathbf{C}^\ell_{dec},\ \mathbf{C}^\ell_{dec} \in \mathbb{R}^{N \times d},
\end{split}
$}
\end{equation}
where $\mathbf{H}^{\ell}_{dec} \in \mathbb{R}^{N \times d}$ denotes the state of the $\ell$-th decoder layer. Then, at each decoding time step $t$, the top-layer ($N_d$-th) decoder hidden state $\mathbf{Z}^{N_d}_{dec,t}$ is fed into the softmax layer to produce the probability distribution of the next target token as:
\begin{equation}
\setlength{\abovedisplayskip}{5pt}
\setlength{\belowdisplayskip}{5pt}
\resizebox{0.89\hsize}{!}{$
\begin{split}
    p(y_{t}|\mathcal{X}_{L_1}, \mathcal{O}, y_{<t}) &= \mathrm{Softmax}(\mathbf{W}_o\mathbf{Z}^{N_d}_{dec,t}+\mathbf{b}_o),\nonumber
\end{split}
$}
\end{equation}
where $\mathbf{W}_o$ and $\mathbf{b}_o$ are trainable weights. 

Finally, the loss function is written as:
\begin{equation}
\setlength{\abovedisplayskip}{5pt}
\setlength{\belowdisplayskip}{5pt}
\begin{split} 
\label{eq:mms}
    \mathcal{L}_{\text{MMS}}^{L_1,L_1} =  -\sum_{t=1}^{N} \log (p (y^{L_1}_t | y^{L_1}_{<t}, \mathcal{X}^{L_1}, \mathcal{V})).
\end{split}
\end{equation}
\textbf{\begin{figure*}[t]
    \centering
    \includegraphics[width=0.94\textwidth]{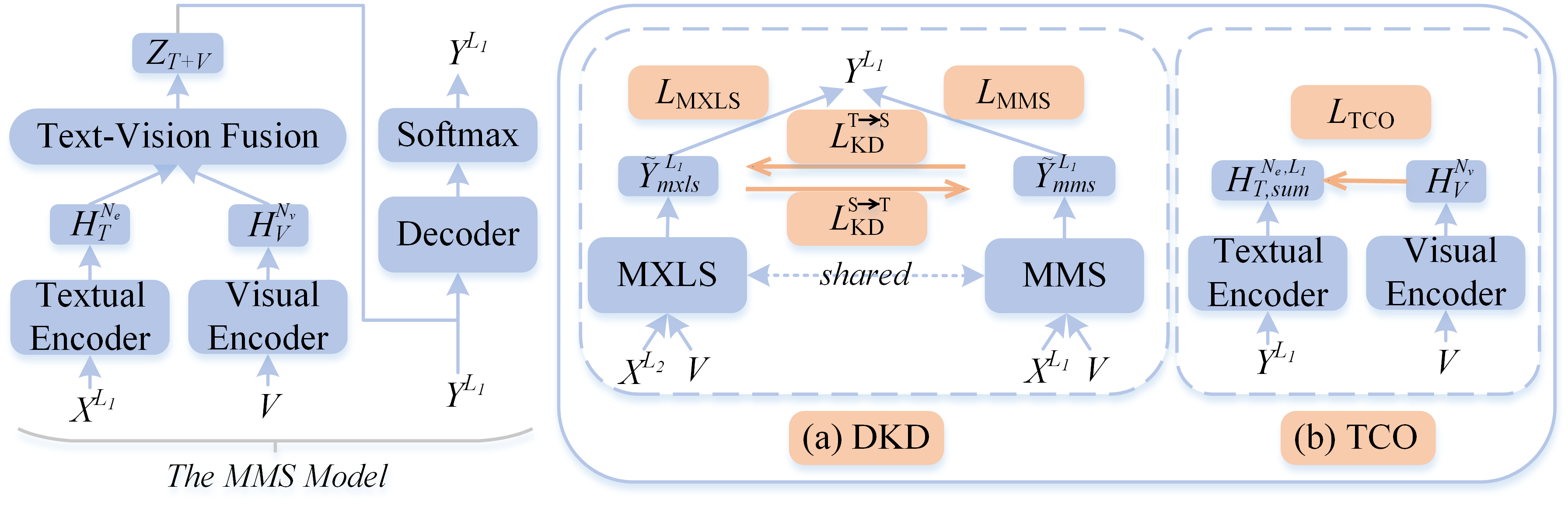}
    \caption{The overview of our model architecture. The left part is a general MMS model, which is enhanced by DKD and TCO. As shown in the right part, the (a) \emph{dual knowledge distillation} (DKD) and (b) \emph{target-oriented contrastive objective} (TCO), are proposed to improve the M$^3$S model performance. }
    \label{fig.2}\vspace{-12pt}
\end{figure*}}

\section{D$^2$TV Training Framework}
\label{sec:sov-mas}
Based on the MMS model described in~\autoref{MMS}, we firstly introduce the proposed \emph{dual knowledge distillation} (DKD) method in~\autoref{sec:vision_features}, which improves both MMS and MXLS tasks. Further, we present a simple yet effective \emph{target-oriented contrastive objective} to filter needless visual information in~\autoref{atask}. Finally, we describe the \emph{training and inference} in~\autoref{training_infer}.

\subsection{Dual Knowledge Distillation}
\label{sec:vision_features}
As shown in the right part of Figure~\ref{fig.2} (a), our framework involves training both MXLS and MMS models. Essentially, the MXLS model needs to simultaneously conduct machine translation and summarization~\cite{liang-etal-2022-variational,wang2022understanding,wang2022survey} while the MMS model only conducts summarization. Obviously, it is harder to train an MXLS model than to learn an MMS model and that is why researchers~\cite{nguyen2022improving,liu-etal-2022-assist} take the MMS model as the teacher to help the MXLS student model (\emph{i.e., teacher$\rightarrow$student} distillation). However, when the MXLS model achieves a level of multilingual and cross-lingual ability, the MXLS model can better transfer and share task knowledge among different languages. Therefore, the MXLS model, in turn, can guide the MMS model to conduct summarization in diverse languages (\emph{e.g.}, English$\rightarrow$English, Indonesian$\rightarrow$Indonesian, Russian$\rightarrow$Russian, and Urdu$\rightarrow$Urdu), especially for low-resource ones (\emph{i.e., student$\rightarrow$teacher} distillation). That is why we propose DKD to mutually enhance their performance. 

\noindent \textbf{Teacher$\rightarrow$Student.} Specifically, for training the student model, given an input $\mathcal{X}^{L_2} = \{x^{L_2}_1, x^{L_2}_2, …, x^{L_2}_{M_1}\}$ in language ${L_2}$ and corresponding visual features $\mathcal{V}$, the student model is to generate the cross-lingual summary $\mathcal{Y}^{L_1} = \{y^{L_1}_1, y^{L_1}_2, …, y^{L_1}_{N}\}$ where $L_2\neq L_1$. Then, we train the student model with two objectives as follows:
\begin{equation}
\label{eq:t2s}
    \mathcal{L}_{\text{student}}^{L_2, L_1} =  \mathcal{L}_{\text{MXLS}}^{L_2, L_1} + \alpha\mathcal{L}_{\text{KD}}^{\text{T}(L_1, L_1)\rightarrow \text{S}(L_2, L_1)},
\end{equation}
where $\mathcal{L}_{\text{MXLS}}^{L_2, L_1}$ denotes by maximizing the likelihood of the ground-truth tokens which takes the cross-entropy form:
\begin{equation}
\label{eq:mxls}
    \mathcal{L}_{\text{MXLS}}^{L_2, L_1} =  -\sum_{t=1}^{N} \log (p (y^{L_1}_t | y^{L_1}_{<t}, \mathcal{X}^{L_2}, \mathcal{V}),
\end{equation}
and $\alpha$ is the trade-off factor and  $\mathcal{L}_{\text{KD}}^{\text{T}(L_1, L_1)\rightarrow \text{S}(L_2, L_1)}$ represents KD loss to penalize the large distance of two hidden states of two summaries generated by the student and teacher models:
\begin{equation}
\label{eq:tkd}
\resizebox{.99\hsize}{!}{%
$
    \mathcal{L}_{\text{KD}}^{\text{T}(L_1, L_1)\rightarrow \text{S}(L_2, L_1)} = \mathrm{dist}(\mathbf{H}_{dec}^{N_d,\text{T}(L_1, L_1)}, \mathbf{H}_{dec}^{N_d,\text{S}(L_2, L_1)}),
$}
\end{equation}
where $\mathrm{dist}(\cdot, \cdot)$ is the distance function to evaluate the difference between two representations (\emph{e.g.}, KL and cosine similarity), and $\mathbf{H}_{dec}^{N_d,\text{T}(L_1, L_1)} = \{\mathbf{h}^T_1,  \mathbf{h}^T_2, …, \mathbf{h}^T_{N}\}$ denote the contextualized representations produced by the decoder of the teacher model, and $\mathbf{H}_{dec}^{N_d,\text{S}(L_2, L_1)} = \{\mathbf{h}^S_1,  \mathbf{h}^S_2, …, \mathbf{h}^S_{N}\}$ denote the representations from the decoder of the student model.


\noindent \textbf{Student$\rightarrow$Teacher.} 
In particular, given the input document $\mathcal{X}^{L_1} = \{x_1^{L_1}, x_2^{L_1}, …, x_{M_2}^{L_1}\}$ in language $L_1$ and corresponding visual features $\mathcal{V}$, the teacher model aims to generate its summary $\mathcal{Y}^{L_1}$ in the same language. We update the parameters of the teacher model with the following objective:
\begin{equation}
\label{eq:s2t}
    \mathcal{L}_{\text{teacher}}^{L_1, L_1} =  \mathcal{L}_{\text{MMS}}^{L_1, L_1} + (1 - \alpha)\mathcal{L}_{\text{KD}}^{\text{S}(L_2, L_1)\rightarrow \text{T}(L_1, L_1)},
\end{equation}
where $\mathcal{L}_{\text{KD}}^{\text{S}(L_2, L_1)\rightarrow \text{T}(L_1, L_1)}$ denotes the inverse KD loss:
\begin{equation}
\label{eq:skd}
\resizebox{.99\hsize}{!}{%
$
    \mathcal{L}_{\text{KD}}^{\text{S}(L_2, L_1)\rightarrow \text{T}(L_1, L_1)} = \mathrm{dist}(\mathbf{H}_{dec}^{N_d,\text{S}(L_2, L_1)}, \mathbf{H}_{dec}^{N_d,\text{T}(L_1, L_1)}).
$}
\end{equation}

Finally, to flexibly distill the knowledge in~\autoref{eq:t2s} and ~\autoref{eq:s2t}, we apply an annealing strategy to dynamically adjust the balancing factor $\alpha$:
\begin{equation}
\label{eq:alpha}
    \alpha = \mathrm{max}(0.5, 1 - t1/T1),
\end{equation}
where $t1$ is the training step ranging from 0 to the max training step $T$ and $T1$ is a hyperparameter. In this manner, the teacher model dominantly guides the student model in the first $T1/2$ training steps and the student model gradually distills the knowledge to the teacher. After training step $T1$, both models begin to equally distill their knowledge to each other.

\subsection{Target-oriented Contrastive Objective}\label{atask}
The M$^3$S task requires a model to have the ability to understand and generate in multiple languages. However, there are some languages that are low-resource and lack enough data to train a good summarizer. Therefore, we aim to take visual features as the bridge between languages and hope that the visual features can be summary-oriented \emph{i.e.}, discarding the noise that not appeared in the summary. To this end, we elaborately design an explicit target-oriented contrastive objective. Particularly, we push the visual feature $\mathcal{V}_i$ close to its corresponding summary $\mathcal{Y}^{L_1}_{i}$ and push apart irrelevant pairs, \emph{e.g.}, ($\mathcal{V}_i$, $\mathcal{Y}_{j}^{L_1}$) where $i \neq j$. Therefore, we treat the paired ($\mathcal{V}_i$, $\mathcal{Y}_{i}^{L_1}$) as the positive sample and treat the pair ($\mathcal{V}_i$, $\mathcal{Y}_{j}^{L_1}$) as the negative samples where $i \neq j$. To obtain the representation of summary and image sequence, we apply \emph{mean-pooling} with mask operation over the summary output $\mathbf{H}^{N_e, L_1}_{T,sum}$ of the $N_e$-th encoder layer and visual output $\mathbf{H}^{N_v}_{V}$ of the $N_v$-th encoder layer, respectively. That is, $\mathbf{h}_{sum}^{L_1}$$=$$\frac{1}{N}\sum_{k=1}^{N}(\mathbf{M}^{sum}_k\mathbf{h}^{N_e,L_1}_{sum,k})$, $\mathbf{h}_{sum}^{L_1} \in \mathbb{R}^{d}$, where $\mathbf{M}^{sum} \in \mathbb{R}^{N}$ denotes the mask matrix, whose value is either 1 or 0 indicating whether the token is padded. Similarly, we obtain the representation of image sequence, \emph{i.e.}, $\mathbf{h}^{vis}$$=$$\frac{1}{m*n}\sum_{i=1}^{n}\sum_{j=1}^{m}(\mathbf{M}^{vis}_{i,j}\mathrm{MLP}(\mathbf{h}^{N_v}_{i,j}))$, $\mathbf{h}^{vis} \in \mathbb{R}^{d}$, where $\mathbf{M}^{vis} \in \mathbb{R}^{n \times m}$ denotes the mask matrix and $\mathrm{MLP}$ is a fully-connected layer. Finally, the target-oriented contrastive training objective is defined by ($B$ is mini-batch size):
\begin{equation}
    \label{eq:tco}
    \begin{aligned}
       \mathcal{L}_{\text{TCO}}^{L_1} = - \log \frac{e^{\mathrm{sim}(\textbf{h}_i^{vis},\ \textbf{h}^{L_1}_{sum, i})/ \tau }}{\sum_{b=1}^Be^\mathrm{sim}({\textbf{h}_i^{vis}, \textbf{h}^{L_1}_{sum, b})/\tau}},
    \end{aligned}
\end{equation}
where $\mathrm{sim}(\cdot, \cdot)$ is the cosine similarity and $\tau$ denotes a temperature hyperparameter. 

\subsection{Training and Inference}\label{training_infer}
At training, we train our model with the following objective:
\begin{equation}
\setlength{\abovedisplayskip}{5pt}
\setlength{\belowdisplayskip}{5pt}
\resizebox{0.99\hsize}{!}{$
\begin{split}
    &\mathcal{J} = \sum_{i=1}^{K}\sum_{j=1,j\neq i}^{K}(\mathcal{L}_{\text{student}}^{L_j, L_i} + \mathcal{L}_{\text{teacher}}^{L_i, L_i} + \beta\mathcal{L}_{\text{TCO}}^{L_i}),
\end{split}\label{eq:final-loss}
$}
\end{equation}

where $K$ is the number of languages and $\beta$ is balancing hyper-parameter. 

Note that the MMS model and the MXLS model are shared and thus the final model can conduct summarization in any language. During inference, the training objectives are not involved and only the model is used to generate summaries.

\begin{table*}[t]
\begin{subtable}{.98\textwidth}
\centering
\resizebox{1.00\textwidth}{!}
{
\begin{tabular}{c|c|c|c|c|c||c}
\hline
\diagbox[dir=NW]{Src}{Trg} & Models & \textbf{English}  & \textbf{Indonesian}    & \textbf{Russian}    & \textbf{Urdu}  & \textbf{Avg.}                                                                                                       \\ \hline
\multirow{5}{*}{\textbf{English} $\rightarrow$}  
& \textbf{MMS}    & \cellcolor{orange!18}{\textbf{36.16 / 13.08} / 27.67 / 70.57} &\cellcolor{blue!12}{\ \;6.87 / \ \;1.94 / \ \;6.34 / 63.39}  &\cellcolor{blue!12}{\ \;1.23 / 0.20 / \ \;1.21 / 59.60} &\cellcolor{blue!12}{\ \;0.14 / \ \;0.00 / \ \;0.14 / 55.44} &\cellcolor{ggray!90}{11.09 / \ \;3.80 / \ \;8.84 / 62.25}    \\
& \textbf{MXLS}    & \cellcolor{orange!18}{\ \;6.94 / \ \;2.35 / \ \;6.03 / 61.82} &\cellcolor{blue!12}{27.23 / \ \;9.32 / 22.13 / 68.40} &\cellcolor{blue!12}{22.52 / 7.88 / 18.07 / 64.84} &\cellcolor{blue!12}{32.27 / 11.17  / 25.15 / 68.29} &\cellcolor{ggray!90}{22.24 / \ \;7.68 / 17.84 / 65.83}    \\
& \textbf{MMS+MXLS} & \cellcolor{orange!18}{35.80 / 13.45 / 27.93 / 70.64} &\cellcolor{blue!12}{27.18 / \ \;9.20 / 22.04 / 68.78} &\cellcolor{blue!12}{23.88 / 8.03 / 19.30 / 65.57} &\cellcolor{blue!12}{28.59 / \ \;8.94 / 22.95 / 66.79} &\cellcolor{ggray!90}{28.86 / \ \;9.90 / 23.07 / 67.94}   \\
& \textbf{Vanilla-KD} & \cellcolor{orange!18}{34.60 / 12.70 / 26.86 / 70.07} &\cellcolor{blue!12}{27.75 / \ \;9.71 / 22.63 / 68.93} &\cellcolor{blue!12}{24.36 / 8.00 / 19.41 / 65.42} &\cellcolor{blue!12}{31.53 / 10.28 / 24.83 / 67.76} &\cellcolor{ggray!90}{29.56 / 10.17 / 23.43 / 68.04}   \\
& \textbf{D$^2$TV  (Ours)}  &\cellcolor{orange!18}{36.12 / 13.21 / \textbf{27.99 / 70.64}} &\cellcolor{blue!12}{\textbf{28.87 / 10.26 / 23.77 / 69.31}} &\cellcolor{blue!12}{\textbf{25.53 / 8.69 / 20.72 / 66.01}} &\cellcolor{blue!12}{\textbf{32.56 / 10.73 / 25.71 / 68.39}} &\cellcolor{ggray!90}{\textbf{30.77 / 10.72 / 24.53 / 68.83}} \\ \hline\hline

\multirow{5}{*}{\textbf{Indonesian} $\rightarrow$}  
& \textbf{MMS}    & \cellcolor{blue!12}{\ \;7.28 / \ \;2.03 / \ \;6.73 / 63.59} &\cellcolor{orange!18}{34.10 / 13.92 / 27.92 / 71.14} &\cellcolor{blue!12}{\ \;1.19 / 0.19 / \ \;1.16 / 60.40}&\cellcolor{blue!12}{\ \;0.13 / \ \;0.01 / \ \;0.13 / 56.42} &\cellcolor{ggray!90}{10.67 / \ \;4.03 / \ \;8.98 / 62.88}    \\
& \textbf{MXLS}    & \cellcolor{blue!12}{32.43 / 11.31 / 25.09 / 69.13} &\cellcolor{orange!18}{\ \;5.39 / \ \;1.59 / \ \;4.88 / 61.82} &\cellcolor{blue!12}{21.65 / 7.80 / 17.62 / 65.09} &\cellcolor{blue!12}{31.85 / 11.00 / 25.39 / 68.53} &\cellcolor{ggray!90}{22.83 / \ \;7.92 / 18.24 / 66.14}   \\
& \textbf{MMS+MXLS} & \cellcolor{blue!12}{32.59 / 11.67 / 25.42 / 69.53} &\cellcolor{orange!18}{\textbf{34.43 / 14.56 / 28.43 / 71.30}} &\cellcolor{blue!12}{24.38 / 8.70 / 20.01 / 66.18} &\cellcolor{blue!12}{30.65 / 10.30 / 24.95 / 67.90} &\cellcolor{ggray!90}{30.51 / 11.31 / 24.70 / 68.72}   \\
& \textbf{Vanilla-KD} & \cellcolor{blue!12}{32.88 / 11.56 / 25.45 / 69.46} &\cellcolor{orange!18}{32.67 / 13.01 / 26.71 / 70.68} &\cellcolor{blue!12}{25.50 / 8.97 / 20.65 / 66.30} &\cellcolor{blue!12}{32.48 / 11.31 / 25.88 / 68.79} &\cellcolor{ggray!90}{30.88 / 11.21 / 24.67 / 68.80} \\
& \textbf{D$^2$TV  (Ours)}  &\cellcolor{blue!12}{\textbf{34.54 / 12.10 / 26.50 / 69.73}} &\cellcolor{orange!18}{33.94 / 14.08 / 28.05 / 71.19} &\cellcolor{blue!12}{\textbf{26.40 / 9.27 / 21.35 / 66.56}} &\cellcolor{blue!12}{\textbf{33.45 / 11.38 / 26.60 / 68.89}} &\cellcolor{ggray!90}{\textbf{32.08 / 11.71 / 25.63 / 69.09}} \\ \hline\hline

\multirow{5}{*}{\textbf{Russian} $\rightarrow$}
& \textbf{MMS}    & \cellcolor{blue!12}{\ \;1.24 / \ \;0.22 / \ \;1.23 / 60.20} &\cellcolor{blue!12}{\ \;1.23 / \ \;0.22 / \ \;1.19 / 60.86}  &\cellcolor{orange!18}{\textbf{30.30 / 11.82} / 24.25 / 68.16} &\cellcolor{blue!12}{\ \;0.11 /\ \; 0.00 / \ \;0.11 / 56.10} &\cellcolor{ggray!90}{\ \;8.22 / \ \;3.02 / \ \;6.69 / 61.33}   \\
& \textbf{MXLS}    & \cellcolor{blue!12}{29.47 / \ \;9.86 / 22.82 / 68.10} &\cellcolor{blue!12}{25.78 / \ \;9.06 / 21.01 / 68.20} &\cellcolor{orange!18}{\ \;2.68 / \ \;0.90 / \ \;2.41 / 59.02} &\cellcolor{blue!12}{31.06 / 10.64 / 25.05 / 68.08}&\cellcolor{ggray!90}{22.24 / \ \;7.61 / 17.82 / 65.85}   \\
& \textbf{MMS+MXLS} & \cellcolor{blue!12}{30.79 / \ \;9.82 / 24.02 / 68.74} &\cellcolor{blue!12}{27.37 / \ \;9.84 / 22.70 / 69.13 }&\cellcolor{orange!18}{29.67 / 11.67 / 24.33 / 68.04} &\cellcolor{blue!12}{30.53 / 10.04 / 24.92 / 67.95} &\cellcolor{ggray!90}{29.59 / 10.37 / 24.09 / 68.50}   \\
& \textbf{Vanilla-KD} & \cellcolor{blue!12}{31.18 / 10.64 / 24.26 / 68.70} &\cellcolor{blue!12}{26.50 / \ \;9.60 / 21.91 / 68.73} &\cellcolor{orange!18}{28.32 / 10.93 / 23.00 / 67.37} &\cellcolor{blue!12}{31.38 / 10.76 / 25.25 / 68.33} &\cellcolor{ggray!90}{29.34 / 10.48 / 23.60 / 68.28}  \\
& \textbf{D$^2$TV (Ours)}  &\cellcolor{blue!12}{\textbf{32.87 / 11.06 / 25.59 / 69.14}} &\cellcolor{blue!12}{\textbf{29.03 / 10.59 / 23.64 / 69.59}}  &\cellcolor{orange!18}{29.90 / 11.44 / \textbf{24.75 / 68.20}} &\cellcolor{blue!12}{\textbf{33.29 / 11.88 / 26.96 / 69.10}} &\cellcolor{ggray!90}{\textbf{31.27 / 11.24 / 25.13 / 68.96}}\\ \hline\hline

\multirow{5}{*}{\textbf{Urdu} $\rightarrow$}  
& \textbf{MMS}    & \cellcolor{blue!12}{\ \;0.09 / 0.00 / \ \;0.09 / 55.58} &\cellcolor{blue!12}{\ \;0.05 / \ \;0.00 / \ \;0.05 / 56.43} &\cellcolor{blue!12}{\ \;0.09 / 0.00 / \ \;0.09 / 56.03}& \cellcolor{orange!18}{37.54 / 15.04 / 30.19 / 70.55}&\cellcolor{ggray!90}{\ \;9.44 / \ \;3.76 / \ \;7.60 / 59.64}  \\
& \textbf{MXLS}    &\cellcolor{blue!12}{29.95 / 9.06 / 23.09 / 68.36}  &\cellcolor{blue!12}{26.00 / \ \;9.37 / 21.44 / 68.43} &\cellcolor{blue!12}{21.52 / 6.87 / 17.23 / 65.35}&\cellcolor{orange!18}{ \;7.70 / \ \;2.62 / \ \;6.15 / 58.76}&\cellcolor{ggray!90}{21.29 / \ \;6.98 / 16.97 / 65.22}    \\
& \textbf{MMS+MXLS} & \cellcolor{blue!12}{28.94 / 9.29 / 22.81 / 67.76 } &\cellcolor{blue!12}{26.43 / \ \;8.84 / 21.74 / 68.55}&\cellcolor{blue!12}{20.47 / 6.44 / 16.74 / 64.33}& \cellcolor{orange!18}{37.72 / 15.78 / 30.97 / 70.96}&\cellcolor{ggray!90}{28.39 / 10.17 / 23.14 / 67.90}    \\
& \textbf{Vanilla-KD} & \cellcolor{blue!12}{29.60 / 9.63 / 23.00 / 67.88}  &\cellcolor{blue!12}{26.30 / \ \;9.02 / 21.65 / 68.29 }&\cellcolor{blue!12}{22.58 / 7.36 / 18.03 / 65.22}& \cellcolor{orange!18}{37.52 / 15.25 / 30.19 / 70.56}&\cellcolor{ggray!90}{29.00 / 10.31 / 23.21 / 67.98}   \\
& \textbf{D$^2$TV  (Ours)}  &\cellcolor{blue!12}{\textbf{32.01 / 9.99 / 24.71 / 68.65}}  &\cellcolor{blue!12}{\textbf{28.23 / 10.01 / 23.19 / 69.25}} &\cellcolor{blue!12}{\textbf{24.52 / 7.87 / 19.98 / 66.13}}&\cellcolor{orange!18}{\textbf{38.05 / 16.12 / 31.30 / 70.97}} &\cellcolor{ggray!90}{\textbf{30.70 / 10.91 / 24.71 / 68.75}}\\ \hline\hline

\end{tabular}
}
\caption{The test results based on the mT5 backbone  in terms of \textsc{ROUGE-1} / \textsc{ROUGE-2} / \textsc{ROUGE-l} / \textsc{BERTScore} scores. } 
\end{subtable}
\begin{subtable}{.980\textwidth}
\centering
\resizebox{1.00\textwidth}{!}
{
\begin{tabular}{c|c|c|c|c|c||c}
\hline
\diagbox[dir=NW]{Src}{Trg} & Models & \textbf{English}  & \textbf{Indonesian}    & \textbf{Russian}    & \textbf{Urdu}  & \textbf{Avg.}                                                                                                       \\ \hline
\multirow{5}{*}{\textbf{English} $\rightarrow$}  
& \textbf{MMS}    & \cellcolor{orange!18}{34.19 / 11.87 / 26.14 / 69.38} &\cellcolor{blue!12}{24.70 / 6.94 / 19.55 / 67.37}  &\cellcolor{blue!12}{21.50 / 6.14 / 17.14 / 64.08} &\cellcolor{blue!12}{23.18 / 5.00 / 17.68 / 63.04} &\cellcolor{ggray!90}{25.89 / 7.49 / 20.13 / 65.96}    \\
& \textbf{MXLS}    & \cellcolor{orange!18}{26.06 / \ \;8.49 /  20.22 / 64.50} &\cellcolor{blue!12}{25.89 / \textbf{8.41} / 21.03 / 66.96} &\cellcolor{blue!12}{\textbf{24.87} / 7.96 /  20.01 / 65.32} &\cellcolor{blue!12}{27.30 / 7.76 /  21.31 / 65.65} &\cellcolor{ggray!90}{25.27 / 8.06 / 20.08 / 65.58}   \\
& \textbf{MMS+MXLS} & \cellcolor{orange!18}{\textbf{35.03} / 12.50 / \textbf{27.17 / 69.75}} &\cellcolor{blue!12}{22.97 / 7.37 / 18.65 / 68.05} &\cellcolor{blue!12}{23.50 / 7.81 / 18.95 / 65.72} & \cellcolor{blue!12}{27.14 / 8.04 / 21.17 / 66.51} &\cellcolor{ggray!90}{27.16 / 8.93 / 21.49 / 67.45}   \\
& \textbf{Vanilla-KD} & \cellcolor{orange!18}{34.84 / \textbf{12.52} / 26.98 / 69.46} &\cellcolor{blue!12}{24.29 / 7.76 / 19.81 / 67.98} &\cellcolor{blue!12}{24.49 / 7.80 / 19.64 / 65.76} &\cellcolor{blue!12}{\textbf{29.06 / 8.83 / 22.84 / 66.90}} &\cellcolor{ggray!90}{28.17 / 9.22 / 22.31 /  67.47}   \\
& \textbf{D$^2$TV  (Ours)}  &\cellcolor{orange!18}{34.78 / 12.36 / 26.81 / 69.55} &\cellcolor{blue!12}{\textbf{26.13} / 8.39 / \textbf{21.15 / 68.26}} &\cellcolor{blue!12}{24.84 / \textbf{8.28 / 20.06 / 65.94}} &\cellcolor{blue!12}{28.60 / 8.44 / 22.30 / 66.70} &\cellcolor{ggray!90}{\textbf{28.59 / 9.37 /  22.58 / 67.71}} \\ \hline\hline

\multirow{5}{*}{\textbf{Indonesian} $\rightarrow$}  
& \textbf{MMS}    &\cellcolor{blue!12}{30.79 / \ \;9.09 / 23.03 / 67.68} & \cellcolor{orange!18}{30.12 / 10.77 /  24.16 / 69.14} &\cellcolor{blue!12}{21.67 / 6.27 /  17.46 / 64.28}&\cellcolor{blue!12}{25.13 / 5.27 / 19.01 / 63.82} &\cellcolor{ggray!90}{26.93 / \ \;7.85 / 20.92 / 66.23}  \\
& \textbf{MXLS}    & \cellcolor{blue!12}{32.85 / 10.56 / 24.78 / 68.53} &\cellcolor{orange!18}{18.27 / \ \;6.45 / 15.00 / 63.64} &\cellcolor{blue!12}{22.58 / 7.29 /  18.07 / 63.65} &\cellcolor{blue!12}{25.96 / 8.10 / 20.70 / 63.36} &\cellcolor{ggray!90}{24.50 / \ \;8.06 / 19.36 / 64.99}    \\
& \textbf{MMS+MXLS} & \cellcolor{blue!12}{33.87 / 11.51 / 26.14 / 69.36} & \cellcolor{orange!18}{29.81 / 11.33 / 24.29 / 69.75} &\cellcolor{blue!12}{24.26 / 8.40 / 19.65 / 65.89} &\cellcolor{blue!12}{29.05 / \textbf{8.99} / 22.87 / 66.88} &\cellcolor{ggray!90}{29.25 / 10.06 / 23.24 / 67.97}  \\
& \textbf{Vanilla-KD} & \cellcolor{blue!12}{33.68 / 11.68 / 25.80 / 69.25} &\cellcolor{orange!18}{30.49 / 11.58 / 24.84 / 69.59} &\cellcolor{blue!12}{24.22 / 8.28 / 19.48 / 66.05} & \cellcolor{blue!12}{29.16 / 8.77 / 23.05 / 67.05} &\cellcolor{ggray!90}{29.38 / 10.07 / 23.29 / 67.98} \\
& \textbf{D$^2$TV  (Ours)}  &\cellcolor{blue!12}{\textbf{34.18 / 11.75 / 26.17 / 69.46}} &\cellcolor{orange!18}{\textbf{31.25 / 11.75 / 25.30 / 69.93}} &\cellcolor{blue!12}{\textbf{24.99 / 8.66 / 20.29 / 66.38}} &\cellcolor{blue!12}{\textbf{29.56} / 8.88 / \textbf{23.18 / 67.28}} &\cellcolor{ggray!90}{\textbf{30.00 / 10.26 / 23.74 / 68.26}} \\ \hline\hline

\multirow{5}{*}{\textbf{Russian} $\rightarrow$}
& \textbf{MMS}    & \cellcolor{blue!12}{29.53 / \ \;8.39 / 22.53 / 66.86} &\cellcolor{blue!12}{24.02 / 6.73 /  19.21 / 66.84}  &\cellcolor{orange!18}{28.29 / \ \;9.85 / 22.33 / 66.88} &\cellcolor{blue!12}{24.79 / 5.16 / 18.92 / 63.59} &\cellcolor{ggray!90}{26.66 / 7.53 / 20.75 / 66.04}   \\
& \textbf{MXLS}    &\cellcolor{blue!12}{32.01 / 10.83 / 24.39 / 67.87} &\cellcolor{blue!12}{23.92 / 8.20 / 19.36 / 65.78}&\cellcolor{orange!18}{23.41 / \ \;8.19 / 18.70 / 62.44} &\cellcolor{blue!12}{24.59 / 7.54 /  19.39 / 63.58} &\cellcolor{ggray!90}{25.23 / 8.47 / 19.93 / 64.91}  \\
& \textbf{MMS+MXLS} & \cellcolor{blue!12}{32.94 / 11.40 / 25.74 / 68.79} &\cellcolor{blue!12}{24.58 / 8.43 / 20.09 / 67.92}&\cellcolor{orange!18}{28.10 / 10.37 / 22.66 / 66.93} &\cellcolor{blue!12}{27.44 / 8.54 / 21.59 / 66.52} &\cellcolor{ggray!90}{28.27 / 9.68 / 22.52 / 67.54}  \\
& \textbf{Vanilla-KD} & \cellcolor{blue!12}{32.86 / 11.54 / 25.64 / 68.75} &\cellcolor{blue!12}{24.63 / 8.40 / 20.15 / 68.12} &\cellcolor{orange!18}{28.27 / 10.31 / 22.64 / 67.04} &\cellcolor{blue!12}{\textbf{28.50 / 8.87} / 22.56 / 66.83} &\cellcolor{ggray!90}{28.56 / 9.78 / 22.75 / 67.69} \\
& \textbf{D$^2$TV  (Ours)}  &\cellcolor{blue!12}{\textbf{33.61 / 11.65 / 26.09 / 69.14}} &\cellcolor{blue!12}{\textbf{26.57 / 8.96 / 21.63 / 68.47}} &\cellcolor{orange!18}{\textbf{28.39 / 10.47 / 22.92 / 67.38}} &\cellcolor{blue!12}{28.39 / 8.81 / \textbf{22.68 / 66.88}} &\cellcolor{ggray!90}{\textbf{29.24 /  9.97 / 23.33 / 67.96}}\\ \hline\hline

\multirow{5}{*}{\textbf{Urdu} $\rightarrow$}  
& \textbf{MMS}    & \cellcolor{blue!12}{24.55 / \ \;5.14 / 18.29 / 64.04} &\cellcolor{blue!12}{19.20 / 4.39 / 15.06 / 64.68} &\cellcolor{blue!12}{17.22 / 3.59 / 13.36 / 61.99}& \cellcolor{orange!18}{34.82 / 12.63 / 27.14 / 68.99}&\cellcolor{ggray!90}{23.95 / \ \;6.44 / 18.46 / 64.92}  \\
& \textbf{MXLS}    & \cellcolor{blue!12}{30.89 / \ \;9.53 / 23.82 / 67.05}  &\cellcolor{blue!12}{22.74 / 7.40 /  18.59 / 65.59} &\cellcolor{blue!12}{21.63 / 6.62 / 17.66 / 63.27}& \cellcolor{orange!18}{21.90 / \ \;6.94 / 17.19 / 60.13}&\cellcolor{ggray!90}{23.69 / \ \;7.49 / 18.84 / 64.01}   \\
& \textbf{MMS+MXLS} & \cellcolor{blue!12}{31.54 / 10.42 / 24.53  / 68.24} &\cellcolor{blue!12}{22.94 / 7.62 / 18.88 / 67.02}&\cellcolor{blue!12}{22.32 / 7.13 / 18.32 / 64.93}& \cellcolor{orange!18}{35.86 / 13.49 / 28.48 / 69.56} &\cellcolor{ggray!90}{28.17 / \ \;9.66 / 22.55 / 67.43}   \\
& \textbf{Vanilla-KD} & \cellcolor{blue!12}{30.96 / 10.20 / 24.34 / 68.18}  &\cellcolor{blue!12}{23.72 / 8.16 / 19.49 / 67.36} &\cellcolor{blue!12}{21.94 / 6.79 / 17.90 / 64.69}&\cellcolor{orange!18}{35.83 / 13.53 / 28.39 / 69.50} &\cellcolor{ggray!90}{28.11 / \ \;9.67 / 22.53 / 67.43}   \\
& \textbf{D$^2$TV  (Ours)}  &\cellcolor{blue!12}{\textbf{31.65 / 10.63 / 24.90 / 68.62}}  &\cellcolor{blue!12}{\textbf{25.47 / 8.52 / 20.68 / 67.75}} &\cellcolor{blue!12}{\textbf{22.38 / 7.19 / 18.44 / 65.37}}&\cellcolor{orange!18}{\textbf{36.46 / 13.76 / 28.75 / 69.94}} &\cellcolor{ggray!90}{\textbf{28.99 / 10.03 / 23.19 / 67.92}}\\ \hline\hline

\end{tabular}
}
\caption{The test results based on the mBART-50 backbone. } 
\end{subtable}
\vspace{-1mm}
\caption{The block in ``\colorbox{orange!18}{{*/*/*/*}}'' denotes the MMS results and the block in ``\colorbox{blue!12}{{*/*/*/*}}'' indicates the MXLS results. The ``\colorbox{ggray!90}{{*/*/*/*}}'' indicates the average (Avg.) score for each model and the best scores in each block are \textbf{bold}. Our \textbf{bold} results indicate that statistically significantly better than the ``Vanilla-KD'' with t-test {\em p} \textless \ 0.05. Note that the results out of each block (e.g., English$\rightarrow$English block) cannot be compared to others (e.g., Indonesia$\rightarrow$English block) because they belong to different language directions. Therefore, in each block of MMS, the MMS always surpasses MXLS without any exception. In each block of MXLS, the MXLS always surpasses MMS without any exception.}
\label{mainres}
\vspace{-5pt}
\end{table*}

\section{Experiments}

\subsection{{\fontfamily{lmtt}\selectfont M$^3$Sum} Dataset} There is no many-to-many multimodal summarization benchmark dataset until now. We construct one as follows. Based on the CrossSum dataset~\cite{bhattacharjee2022crosssum} and MM-Sum dataset~\cite{https://doi.org/10.48550/arxiv.2212.07672}, we construct a \textbf{M}any-to-\textbf{M}any \textbf{M}ultimodal \textbf{Sum}marization ({\fontfamily{lmtt}\selectfont M$^3$Sum}) dataset. The original CrossSum dataset is crawled from the BBC website\footnote{\url{https://www.bbc.com/}} and its quality has been verified and ensured reliability by~\citet{bhattacharjee2022crosssum}. However, the lack of associated image sequence in CrossSum, makes it impossible to directly conduct research on MMS and MXLS. The original MM-Sum dataset is also crawled from the BBC website, which includes multilingual multimodal summarization. But it cannot conduct cross-lingual summarization due to the lacking of cross-lingual alignment. Therefore, we reorganize both datasets and conduct cross-lingual alignment through the same $url$ in each dataset.

According to the dataset size of each language, we follow CrossSum~\cite{bhattacharjee2022crosssum} and utilize about 80\% training:10\% validation:10\% test splitting. Besides, in CrossSum, the number of languages is 44 and thus there are 44*44 language directions. \autoref{tab:dataset} of~\autoref{data-appendix} shows the detailed statistic of our {\fontfamily{lmtt}\selectfont M$^3$Sum} and please refer to it for details.

\subsection{Setup and Metrics}
\label{sect:data}

\noindent \textbf{Implementation Details.}
For efficiency, we randomly select 4 languages (\emph{i.e.}, English, Indonesian, Russian, and Urdu), which totally cover 16 language directions. Please refer to~\autoref{ID} for more implementation details including data pre-processing and hyper-parameters settings.

\noindent \textbf{Metrics.}
Following~\citet{bhattacharjee2022crosssum,wang2023cross}, we use the standard ROUGE scores (\textsc{ROUGE-1}, \textsc{ROUGE-2}, and \textsc{ROUGE-L})~\cite{lin-2004-rouge} with the statistical significance test~\cite{koehn-2004-statistical} for a fair comparison. Besides, we apply \textsc{BERTScore}~\cite{bert-score} for a comprehensive comparison.

\subsection{Comparison Models}
\label{ssec:layout}
\begin{itemize}[leftmargin=*]
\item \textbf{MMS}: It is the MMS model trained with the objective~\autoref{eq:mms}.
\item \textbf{MXLS}: It is the MXLS model trained with the objective~\autoref{eq:mxls}.
\item \textbf{MMS+MXLS}: It is the model jointly trained with the objectives~\autoref{eq:mms} and~\autoref{eq:mxls}, which actually is the M$^3$S training objective. 
\item \textbf{Vanilla-KD}: It is the model enhanced with the knowledge distillation, which is trained with the objectives ~\autoref{eq:mms},~\autoref{eq:mxls} and~\autoref{eq:t2s}.
\item \textbf{D$^2$TV }: It is the proposed model which are trained with the objective~\autoref{eq:final-loss}.

\end{itemize}

All the above models use the multimodal Transformer described in~\autoref{MMS} and involve two strong training backbones: \textbf{mT5}~\cite{xue-etal-2021-mt5} and \textbf{mBART-50}~\cite{tang-etal-2021-multilingual}. 

\begin{table*}[t!]
\centering
\scalebox{0.62}{
\begin{tabular}{l|l|c|c|c|c||c}
\hline
&\textbf{Models} &\textbf{English$\rightarrow$*}&\textbf{Indonesian$\rightarrow$*} &\textbf{Russian$\rightarrow$*} &\textbf{Urdu$\rightarrow$*}&\textbf{Train (S)} \\
\hline
0& MMS+MXLS &\cellcolor{ggray!90}{28.86 / \ \;9.90 / 23.07 / 67.94}  &\cellcolor{ggray!90}{30.51 / 11.31 / 24.70 / 68.72} &\cellcolor{ggray!90}{29.59 / 10.37 / 24.09 / 68.50}&\cellcolor{ggray!90}{28.39 / 10.17 / 23.14 / 67.90} &\cellcolor{green!15}{\ \;6.12} \\
1&\emph{w/o} Visual Features &\cellcolor{ggray!90}{28.48 / \ \;9.44 / 22.73 / 67.71  }&\cellcolor{ggray!90}{30.12 / 10.83 / 24.33 / 68.39 }&\cellcolor{ggray!90}{29.18 / 10.01 / 23.68 / 68.33}&\cellcolor{ggray!90}{27.91 / \ \;9.88 / 22.79 / 67.67} &\cellcolor{green!15}{\ \;5.37}\\\cdashline{1-6}[4pt/2pt]
2&\emph{w/} Vanilla KD &\cellcolor{ggray!90}{29.56 / 10.17 / 23.43 / 68.04 }&\cellcolor{ggray!90}{30.88 / 11.21 / 24.67 / 68.80}&\cellcolor{ggray!90}{29.34 / 10.48 / 23.60 / 68.28}&\cellcolor{ggray!90}{29.00 / 10.31 / 23.21 / 67.98} &\cellcolor{green!15}{\ \;7.95} \\
3&\emph{w/} DKD &\cellcolor{ggray!90}{30.19 / 10.63 / 23.98 / 68.48 }&\cellcolor{ggray!90}{31.49 / 11.44 / 25.12 / 68.92}&\cellcolor{ggray!90}{30.89 / 10.88 / 24.51 / 68.55}&\cellcolor{ggray!90}{29.96 / 10.65 / 24.38 / 68.51}  &\cellcolor{green!15}{\ \;9.58}\\\cdashline{1-6}[4pt/2pt]
4&\emph{w/} CAT &\cellcolor{ggray!90}{30.19 / 10.57 / 24.03 / 68.45 }&\cellcolor{ggray!90}{31.42 / 11.49 / 25.33 / 68.88}&\cellcolor{ggray!90}{30.75 / 10.59 / 24.41 / 68.64}&\cellcolor{ggray!90}{29.96 / 10.61 / 24.32 / 68.31}  &\cellcolor{green!15}{13.45}\\
5&\emph{w/} TCO &\cellcolor{ggray!90}{30.01 / 10.45 / 23.77 / 68.27 }&\cellcolor{ggray!90}{31.27 / 11.25 / 25.01 / 68.85}&\cellcolor{ggray!90}{30.78 / 10.61 / 24.45 / 68.51}&\cellcolor{ggray!90}{29.89 / 10.44 / 24.15 / 68.29} &\cellcolor{green!15}{\ \;8.90} \\
\cdashline{1-6}[4pt/2pt]
6&\emph{w/} DKD\&TCO &\cellcolor{ggray!90}{\textbf{30.77 / 10.72 / 24.53 / 68.83} }&\cellcolor{ggray!90}{\textbf{32.08 / 11.71 / 25.63 / 69.09}}&\cellcolor{ggray!90}{\textbf{31.27 / 11.24 / 25.13 / 68.96}}&\cellcolor{ggray!90}{\textbf{30.70 / 10.91 / 24.71 / 68.75}} &\cellcolor{green!15}{11.88} \\
\hline
\end{tabular}}
\caption{Ablation study based on the mT5 (\colorbox{ggray!90}{Avg.} results of \textsc{ROUGE-1} / \textsc{ROUGE-2} / \textsc{ROUGE-l} / \textsc{BERTScore}), where each component is separately added on the ``MMS+MXLS''. ``*'' denotes the four languages (\emph{i.e.}, English, Indonesian, Russian, and Urdu). The ``CAT'' denotes the complex auxiliary tasks of~\cite{https://doi.org/10.48550/arxiv.2212.07672}. \colorbox{green!12}{Train (S)} denotes how many seconds are required for each model to train one step (32 batch size * 8 GPUs).}\label{tab:abl}\vspace{-12pt}
\end{table*}

\subsection{Main Results}
\autoref{mainres} presents the main results on many-to-many scenarios grounding on different backbones. Overall, our model obtains significantly better results than all contrast models in both settings. 

\noindent \textbf{Results based on mT5 backbone.} 
\label{ssec:ende}
In \autoref{mainres} (a), 1) in each group (\emph{e.g.}, English$\rightarrow$\{English, Indonesian, Russian, Urdu\}), the MMS model typically performs better in generating monolingual summaries while it cannot process well in cross-lingual settings. The reason is that the MMS model has no access to the cross-lingual data during training. The MXLS model faces a similar phenomenon where it cannot handle well the monolingual summaries while generating better cross-lingual summaries. In contrast, the ``MMS+MXLS'' model, as a multitask model, achieves better results than both MMS and MXLS models, showing that the MMS and MXLS tasks can benefit each other and thus improve both of them. Based on this finding, a dual knowledge distillation is more reasonable than unidirectional knowledge distillation. Our results further demonstrate this (See ablation study). 2) Generally, in each block, we find that our D$^2$TV  approach notably outperforms the Vanilla-KD method, showing the effectiveness of dual knowledge distillation and target-oriented contrastive learning. Although our results are slightly worse than the MMS model in English$\rightarrow$English, Indonesian$\rightarrow$Indonesian, and Russian$\rightarrow$Russian directions of ``\colorbox{orange!18}{{*/*/*/*}}'' blocks, our D$^2$TV  model can balance well between MMS and MXLS. The results in each ``\colorbox{ggray!90}{Avg.}'' blocks fully prove this point. 3) On average, our model consistently and significantly surpasses all baselines by large margins (\emph{e.g.}, the previous best ``Vanilla-KD'', up to \textbf{1.70/0.60/1.50} ROUGE and \textbf{0.77} BERTScore scores in Urdu$\rightarrow$* directions, respectively).

\noindent \textbf{Results based on mBART-50 backbone.}
\label{ssec:chen}
In~\autoref{mainres} (b), we observe similar findings as in the mT5-based scenario. This demonstrates that our conclusions are solid and convincing on general pre-trained language models. All these results prove the superiority of our approach.

\section{Analysis}
\subsection{Ablation Study}
\label{ssec:abs}
We conduct ablation studies to investigate how well each component works. The results are shown in~\autoref{tab:abl}. We have the following conclusions: 
\begin{itemize}[leftmargin=*]
\item (Row 1 vs. row 0). The results show that incorporating visual features has a positive impact on the model performance, demonstrating the importance of image sequence for the summary. 
\item (Row 2 vs. row 0). The vanilla KD makes reasonable contributions, showing that the MMS model indeed helps improve the quality of summaries in terms of both ROUGE and BERTScore scores, suggesting that distilling the knowledge of MMS to MXLS is helpful to summarization; 
\item (Row 3 vs. row 2\&row 0). The results show that dual knowledge distillation further improves the model performance, indicating that the knowledge of MMS and MXLS are beneficial to each other and thus can enhance both of them.
\item (Row 5 vs. row 4\&row 0). The results show that summary-oriented visual features can significantly improve the quality of summaries and our simple TCO  achieves comparable performance with the CAT with less training time. This shows the superiority of the target-oriented contrastive objective. 
\item (Row 6 vs. row 0). Adding DKD and TCO exhibit notable cumulative benefits, showing the effectiveness of the proposed approach. 
\end{itemize}


\subsection{Human Evaluation}
Following~\citet{https://doi.org/10.48550/arxiv.2212.07672}, we conduct human studies on 50 samples randomly selected from English$\rightarrow$English and Russian$\rightarrow$English test sets to further evaluate the performance of all models. We invite three Chinese postgraduate students who major in English to compare the generated summaries~\footnote{When evaluating summaries in Russian$\rightarrow$English, we show them the English document rather than the Russian document where the English and Russian document describe the same thing.} and assess each summary from three independent aspects: \textbf{fluency} (Flu.), \textbf{conciseness} (Con.) and \textbf{informativeness} (Inf.). We ask them to score each aspect from 1 (worst) to 5 (best). The average results are presented in~\autoref{human1}. 

\autoref{human1} shows the human results. We find that our D$^2$TV  substantially outperforms all contrast models under all criteria in both directions, which further shows the effectiveness and superiority of our approach. The Fleiss' Kappa scores~\cite{doi:10.1177/001316447303300309} of Flu., Con. and Inf. are 0.74, 0.70 and 0.65, respectively, which indicates a substantial agreement among three evaluators. Furthermore, we present a case study in~\autoref{CS} and it intuitively shows the superiority of our D$^2$TV.

\begin{table}[]
\small
\centering
\scalebox{0.9}{
\begin{tabular}{@{}l|ccc|ccc@{}}
\hline
\multirow{2}{*}{\textbf{Models}} & \multicolumn{3}{c|}{\textbf{English}$\rightarrow$\textbf{English}} & \multicolumn{3}{c}{\textbf{Russian}$\rightarrow$\textbf{English}} \\ \cline{2-7} 
 & Flu. & Con. & Inf. & Flu. & Con. & Inf. \\ 
\hline
MMS & 3.28 & 3.04 & 2.58 & 1.20 & 1.04 & 0.88 \\
MXLS & 1.22 & 1.10 & 0.96 & 2.72 & 2.28 & 2.14 \\
MMS+MXLS & 3.44 & 3.28 & 3.18 & 3.26 & 3.16 & 3.04 \\
Vanilla-KD & 3.60 & 3.46 & 3.22 & 3.40 & 3.28 & 3.18 \\
D$^2$TV  & \textbf{4.14} & \textbf{3.94} & \textbf{3.78} & \textbf{3.86} & \textbf{3.58} & \textbf{3.50} \\ 
\hline
\end{tabular}}
\caption{Human evaluation results. `Flu.': fluency, `Con.': conciseness, and `Inf.': informativeness.}
\label{human1}\vspace{-12pt}
\end{table}


\section{Related Work}

\noindent \textbf{Multimodal Monolingual Summarization (MMS).} 
With the rapid growth of multimedia, many MMS datasets have been built which cover video summarization~\cite{5711541,sanabria18how2}, movie summarization~\cite{6527322}, meeting records summarization~\cite{1221239}, sentence summarization~\cite{li2018multi,li-etal-2017-multi}, product summarization~\cite{li2020aspect}, and news summarization~\cite{zhu-etal-2018-msmo,chen-zhuge-2018-abstractive,hasan-etal-2021-xl,fu-etal-2021-mm,https://doi.org/10.48550/arxiv.2212.07672}. With the data resources extensively used, the MMS task has attracted much attention, where the existing work mainly focuses on 1) how to effectively exploit the additional features which are generally implicitly learned by the MMS objective or 2) explicit and complex auxiliary tasks, having achieved impressive performance on these high-resource English datasets~\cite{li2018read,li-etal-2020-vmsmo,zhu2020multimodal,zhu2021graph,zhang2021unims,zhang2021hierarchical,yu-etal-2021-vision}. In this work, we instead of focusing on introducing a more general and practical many-to-many multimoal summarization setting and also provide a corresponding benchmark dataset. Additionally, we propose a simple yet effective target-oriented contrastive learning objective to filter needless visual features, \emph{i.e.}, offer summary-oriented visual features.

\noindent \textbf{Multimodal Cross-lingual Summarization (MXLS).} 
There is only one study that focuses on the MXLS task,  \emph{i.e.},~\citet{liu-etal-2022-assist} first propose this task and design a triple-stage training framework and distill the knowledge from MMS to enhance MXLS while ignoring the performance of MMS. Different from this work, we introduce the many-to-many multimodal summarization task. Furthermore, we devise a dual knowledge distillation approach to simultaneously improve both MMS and MXLS tasks. 

\noindent \textbf{Knowledge Distillation (KD).} 
KD~\cite{ori_kd} is to transfer the knowledge (\emph{e.g.}, soft targets outputs) of the stronger model (aka. the teacher model) to the small model (aka. the student model), which has achieved impressive results in the literature~
\cite{zhang2023towards}. In summarization, ~\cite{zhang2021unims} adopt KD from a vision-language pre-trained model to improve image selection when generating multimodal summaries. Besides, researchers~\cite{nguyen2022improving,liu-etal-2022-assist} typically treat the monolingual summarization model as the teacher model and the cross-lingual one as the student model because the monolingual summarization model is easier to train well than the cross-lingual one, which has shown promising performance on cross-lingual summarization task while ignoring the performance of the monolingual one. In this work, we aim to mutually prompt both monolingual and cross-lingual summarization tasks via dual KD rather than only improving the cross-lingual summarization task by unidirectional KD.

\noindent \textbf{Constrastive Learning.} 
The idea of contrastive learning aims to learn effective representation by pulling semantically close neighbors together and pushing apart non-neighbors~\cite{1640964}, which has verified its superiority in many fields~\cite{zhou2023rc3}. In summarization, \citet{liu-liu-2021-simcls} use contrastive loss to post-rank generated summaries and achieves good results in textual-only benchmark datasets. \citet{cao-wang-2021-cliff} and~\citet{xu2021sequence} use contrastive learning to improve faithfulness and factuality and observe consistent improvements.~\citet{wang-etal-2021-contrastive} apply contrastive learning for multilingual summarization and obtain promising performance. Differently, we introduce it into the multimodal area and aim to pull the visual feature close to its corresponding summary and offer summary-oriented visual features. Therefore, we can improve the quality of summaries from the perspective of visual features rather than the textual document.

\section{Conclusion}
In this paper, we first introduce a more general task, \emph{i.e.}, M$^3$S, which can support both MMS and MXLS tasks. Further, we propose a dual knowledge distillation and target-oriented vision (D$^2$TV) enhanced framework for the new task. Extensive experiments demonstrate that our model significantly outperforms related baselines in terms of ROUGE, BERTScore scores, and human evaluation. Furthermore, we contribute a many-to-many multimodal summarization ({\fontfamily{lmtt}\selectfont M$^3$Sum}) dataset to the research community. 

\section*{Limitations}
Although we show that our D$^2$TV  outperforms the vanilla-kD model based on two stronger backbone \emph{i.e.}, mT5~\cite{xue-etal-2021-mt5} and mBART-50~\cite{tang-etal-2021-multilingual}, there are some limitations worth considering to study in future work: (1) In this study, we only provide 44 languages and conduct experiments on four out of them, and future work could extend our method to more languages; (2) With the development of the large-scale language models, extending and validating our approach on them may be future work. 

\section*{Ethics Statement}
In this section, we consider the potential ethical issues of our model. In this paper, we propose D$^2$TV  which is trained on the publicly-available BBC datasets. Therefore, D$^2$TV might lead to incorrect summaries in applications and involve the same biases and toxic behaviors exhibited by the datasets. Besides, we obtained our {\fontfamily{lmtt}\selectfont M$^3$Sum} dataset by reorganizing the CrossSum~\cite{bhattacharjee2022crosssum} and MMSum~\cite{https://doi.org/10.48550/arxiv.2212.07672} datasets\footnote{The data originally comes from: https://www.bbc.com/} and its permissions are granted to copy, distribute and modify the contents under the terms of the \href{https://en.wikipedia.org/wiki/Wikipedia:Text_of_Creative_Commons_Attribution-ShareAlike_3.0_Unported_License}{Creative Commons AttributionShareAlike 3.0 Unported License} and \href{https://www.wikidata.org/wiki/Wikidata:Text_of_the_Creative_Commons_Public_Domain_Dedication}{Creative Commons CC0 License}, respectively.

\section*{Acknowledgements}
The research work described in this paper has been supported by the National Key R\&D Program of China (2020AAA0108001)and the National Nature Science Foundation of China (No. 61976015, 61976016, 61876198 and  61370130). The authors would like to thank the anonymous reviewers for their valuable comments and suggestions to improve this paper.

\bibliography{anthology,custom}
\bibliographystyle{acl_natbib}

\clearpage
\newpage
\appendix
\section{Dataset Statistics and Splits.}
\label{data-appendix}
As shown in \autoref{tab:dataset}, we only present 4*4 language directions of our {\fontfamily{lmtt}\selectfont M$^3$Sum} used in this work for simplicity. Actually, our {\fontfamily{lmtt}\selectfont M$^3$Sum} covers 44*44 languages and in total includes 1,078,215 article-summary pairs with 3,479,348 images, where each article-summary pair contains about 3.23 images on average. The average article and summary length for all languages is about 520 and 84, respectively. According to the dataset size of each language, we follow CrossSum~\cite{bhattacharjee2022crosssum} and utilize about 80\% training:10\% validation:10\% test splitting. Besides, in CrossSum, the number of languages is 44 and thus there are 44*44 language directions. For efficiency, we randomly select 4 languages (\emph{i.e.}, English, Indonesian, Russian, and Urdu), which totally cover 16 language directions. 

\begin{table}[h]
\centering
\setlength{\tabcolsep}{0.5mm}{
\begin{tabular}{lrrrr}
\hline
\textbf{Languages} &\textbf{English}&\textbf{Indonesian} &\textbf{Russian} &\textbf{Urdu} \\
\hline
\textbf{English} &24,768&10,037 &9,076 &6,297 \\
\textbf{Indonesian} &9,814&23,176 &7,260 &6,324 \\
\textbf{Russian} &8,902&7,329 &21,036 &5,179 \\
\textbf{Urdu} &6,052&5,810 &4,700 &17,800 \\
\hline
\end{tabular}}
\caption{An example of 4 * 4  Language pairs covered by our M$^3$Sum dataset. }\label{tab:dataset}
\end{table}

\section{Implementation Details}
\label{ID}
\noindent \textbf{Data Pre-Processing.} Following~\citet{bhattacharjee2022crosssum}, we pre-process the textual data by truncating or padding them into sequences of 512 tokens for $\mathcal{X}$ and the outputs $\mathcal{Y}$ to 84 tokens after using the 250k wordpiece~\cite{xue-etal-2021-mt5} vocabulary provided with the mT5 checkpoint (similar to mBART-50 setting). For the image sequence, following~\citet{https://doi.org/10.48550/arxiv.2212.07672}, we truncate or pad the sequence length to 180 (\emph{i.e.}, five images: 5 * 36; n=5, m=36).

\noindent \textbf{Hyper-Parameters.} In this work, we use two strong backbones, \emph{i.e.}, the $base$\footnote{\url{https://huggingface.co/google/mt5-base/tree/main}} model of mT5~\cite{xue-etal-2021-mt5} and the $large$\footnote{\url{https://huggingface.co/facebook/mbart-large-50-many-to-many-mmt}} model of mBART-50~\cite{tang-etal-2021-multilingual}. We list detailed hyper-parameter used in this work in~\autoref{tab:train-detail}. 

For inference, we use beam search with beam size 4 and length penalty of $\gamma$ = 0.6. When calculating the ROUGE scores, we use the multi-lingual rouge\footnote{\url{https://github.com/csebuetnlp/xl-sum/tree/master/multilingual_rouge_scoring}} toolkit following~\citet{bhattacharjee2022crosssum}. All experimental results reported in this paper are the average of three runs with different random seeds. 

\begin{table}[]
    \centering
    \resizebox{\linewidth}{!}{
    \begin{tabular}{l|c|c}
        \hline
        \multirow{1}{*}{\textbf{Hyperparameters}} & \multicolumn{1}{c|}{\textbf{mT5}}  & \multicolumn{1}{c}{\textbf{mBART-50}} \\
        \hline
        batch size ($B$) & 32 & 32 \\
        number of GPUs & 8 V100 & 8 V100 \\
        hidden size & 768 & 1024 \\
        filter size & 2048 & 4096  \\
        encoder layers & 12 & 12 \\
        decoder layers & 12 & 12  \\
        attention heads & 12 & 16 \\
        label smoothing & 0.1 & 0.1  \\
        learning rate & 2e-5 & 5e-6  \\
        warmup steps & 2,000 & 2,000  \\
        training steps $T$ &10,000 &10,000\\
        $T1$ &5,000 &5,000\\
        training time &$\approx$33h &$\approx$8h\\
        optimizer &Adam &Adam\\
        adam beta1 &0.9 &0.9\\
        adam beta2 &0.998 &0.98\\
        layer normalization & postnorm & postnorm \\
        $M$ &520 &520\\
        $N$ &84 &84\\
        $m$ &5 & 5\\
        $n$ &36 & 36\\
        $N_e$ &12 & 12\\
        $N_d$ &12 & 12\\
        $N_v$ &4 & 4\\
        $d$ &768 & 1024\\
        $d_v$ &2048 & 2048\\
        $d_c$ &256 & 256\\
        $K$ &4&4 \\
        $\beta$ &1.0 &1.0\\
        \hline
    \end{tabular}
    }
    \caption{Training hyperparameters and model configurations of our experiments.}
    \label{tab:train-detail}
\end{table}

\section{Case Study}
\label{CS}
\autoref{cs} shows an example of the many-to-many multimodal summarization, the generated summary, and the ground truth summary in different languages. (updating later.)

\textbf{\begin{figure*}[t]
    \centering
    \includegraphics[width=0.98\textwidth]{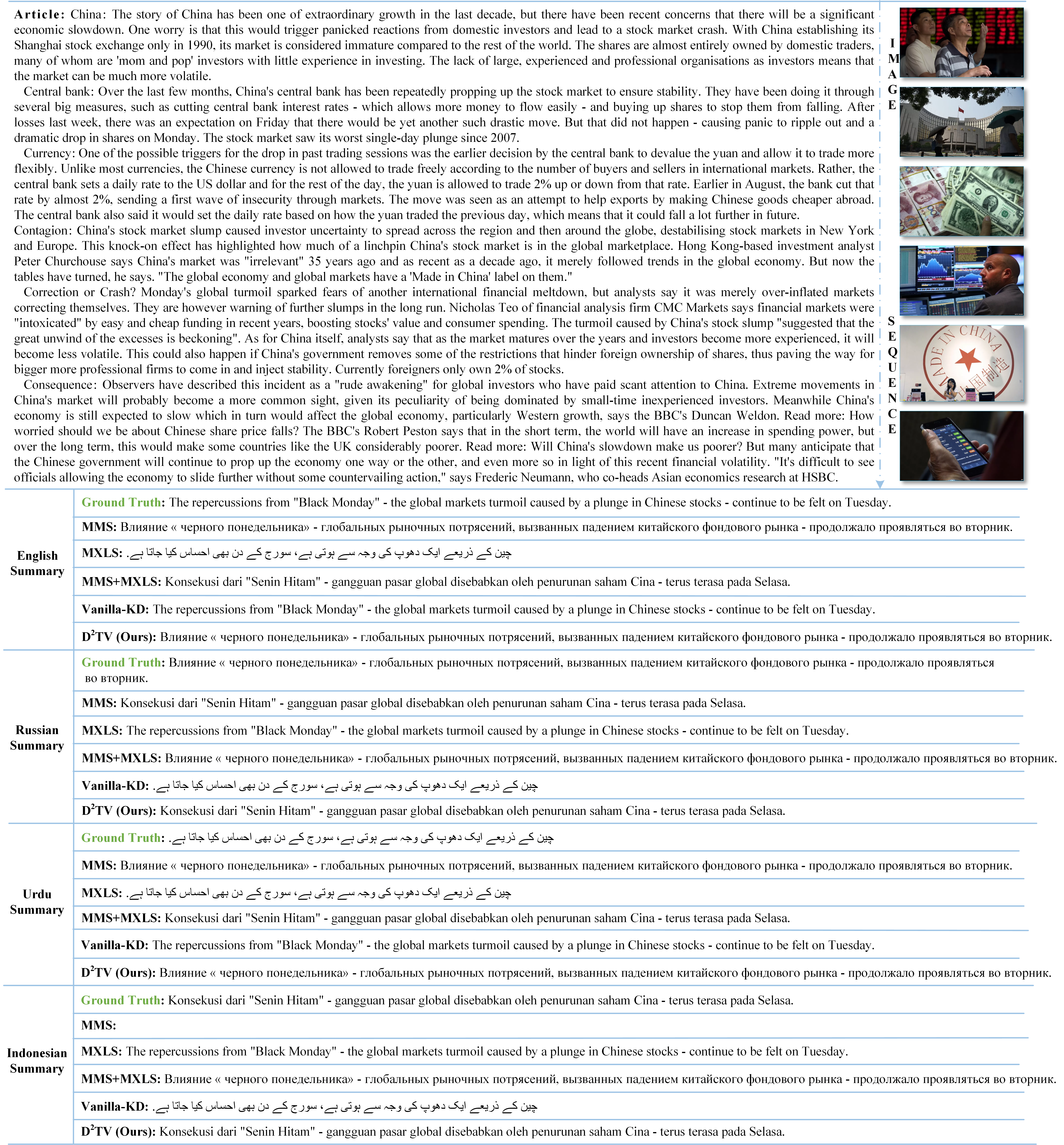}
    \caption{An example of many-to-many summarization in different language directions. ((updating later.))}
    \label{cs}
\end{figure*}}



\end{document}